\documentclass[conference]{IEEEtran}

\usepackage[utf8x]{inputenc}
\usepackage[T1]{fontenc}
\usepackage{graphicx}
 \usepackage[colorlinks=true, linkcolor=blue, urlcolor=magenta, citecolor=green]{hyperref}
\usepackage[english]{babel}
\usepackage{algorithm}
\usepackage[noend]{algpseudocode}
\usepackage{amsmath,amssymb,amsfonts}
\usepackage[numbers]{natbib}
\usepackage{multirow}
\usepackage{multicol}
\usepackage{caption}
\usepackage{subcaption}
\newcommand{\R}{\mathbb{R}}
\usepackage{float}

\usepackage{url}

\usepackage{breakurl}

\begin{document}

\title{FGDCC: Fine-Grained Deep Cluster Categorization - A Framework for Intra-Class Variability Problems in Plant Classification}

\author{\IEEEauthorblockN{Luciano Araújo Dourado Filho}
    \IEEEauthorblockA{
        \textit{Universidade Estadual de Feira de Santana}\\
        Feira de Santana, Brasil\\
        lucianoadfilho@ecomp.uefs.br
    }
    \and
    \IEEEauthorblockN{Rodrigo Tripodi Calumby}
    \IEEEauthorblockA{
        \textit{Universidade Estadual de Feira de Santana}\\
        Feira de Santana, Brasil\\
        rtcalumby@uefs.br
    }
}

\maketitle

\thispagestyle{plain}
\pagestyle{plain}

\begin{abstract}
Intra-class variability is given according to the significance in the degree of dissimilarity between images within a class. In that sense, depending on its intensity, intra-class variability can hinder the learning process for DL models, specially when such classes are also underrepresented, which is a very common scenario in Fine-Grained Visual Categorization (FGVC) tasks. This paper proposes a novel method that aims at leveraging classification performance in FGVC tasks by learning fine-grained features via classification of class-wise cluster assignments. Our goal is to apply clustering over each class individually, which can allow to discover pseudo-labels that encodes a latent degree of similarity between images. In turn, those labels can be employed in a hierarchical classification process that allows to learn more fine-grained visual features and thereby mitigating intra-class variability issues. Initial experiments over the PlantNet300k enabled to shed light upon several key points in which future work will have to be developed in order to find more conclusive evidence regarding the effectiveness of our method. Our method still achieves state-of-the-art performance on the PlantNet300k dataset even though some of its components haven't been shown to be fully optimized. Our code is available at \href{https://github.com/ADAM-UEFS/FGDCC}{https://github.com/ADAM-UEFS/FGDCC}.
\end{abstract}

\renewcommand\IEEEkeywordsname{Keywords}

\IEEEpeerreviewmaketitle

\section{Introduction}
Learning visual characteristics from high intra-class variability data can be a challenging task for Deep Learning (DL) models, especially when the datasets have long-tailed distributions. Problems like these are oftenly associated to Fine-Grained Visual Categorization (FGVC)~\cite{basirat_relu}. In contrast to Coarse-Grained Visual Categorization (CGVC), in FGVC, the goal is to discriminate between distinct classes from the same category (e.g., dog breeds, plant diseases, car brands, bird species,  etc). FGVC is inherently challenging because in order to discriminate between similar objects (i.e., inter-class similarity) the models have to account for the minor details in the images, which sometimes can be even harder considering lower resolutions.

Another classical example of a challenging FGVC task is plant species recognition. Many plant organisms from different species share common structures (e.g., leaves, flowers, stem, etc), which makes feature overlap between different species very common. Despite inter-class similarity, often times there is also the presence of many underrepresented categories leading to long-tailed distributions and therefore, generalization issues. Besides that, plants are intrinsically prone to intra-class variability as well~\cite{feitoza2019hibrido}. That is because plant individuals commonly suffer the action of predators, diseases, influence of climatic factors, aging, seasonality, etc., all of which can affect its structure and modify their visual appearance. Because of that, visual dissimilarity between morphological characteristics of specimens from the same species is somewhat expected. In such scenarios, a model can fail to account for common patterns present in the data, which may prevent it from learning a generalizable representation. Besides that, many cases of plant species recognition involve multi-organ classification scenarios, in which the dataset classes are composed by images of different views -- different organs (i.e., leaves, flowers, fruits) with very distinct morphological features from each other. 




\begin{figure}[t]
    \centering
    \includegraphics[width=\columnwidth]{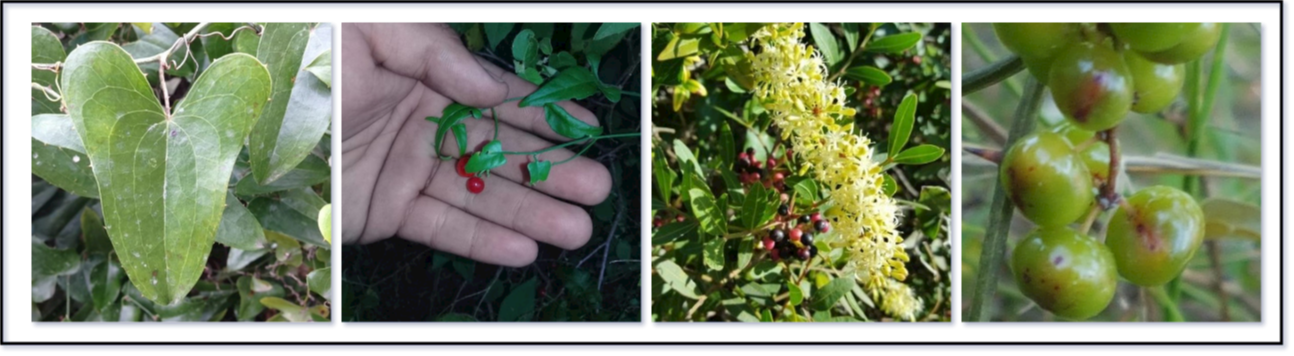}
    \caption{Example of intra-class variability for the species Smilax aspera L.}
    \label{fig:enter-label}
    \vspace{-0.5cm}
\end{figure}

Recent works have demonstrated that combining supervised and unsupervised approaches could be helpful in leveraging classification performance in Deep Neural Networks (DNN's) and minimizing the impact of the variability in the learning process~\cite{feitoza2019hibrido, caron2019deeper}. In the work of~\cite{feitoza2019hibrido}, for example, the authors proposed a framework that combines supervised and unsupervised learning to address intra-class variability in plant species recognition tasks. Their idea consists of individually clustering the images that composes each species into subgroups that will replace the pre-existing classes. Through that, the method is capable of redefining the original classes configuration into a new setting of labels that is expected to be composed by more homogeneous samples and therefore to ease the learning process. Despite that, in the authors framework, clustering and feature extraction are performed as sequential steps of different processes, which fails to leverage the advantages of Deep Neural Networks (DNN) in terms of learning capabilities. On the other hand, in~\cite{caron2019deeper}, the authors propose a similar clustering framework but as an end-to-end approach for self-supervised pre-training of DNN's on non-curated data. However, the authors framework was applied in a Self-Supervised Learning (SSL) context, with a primitive Convolutional Neural Network (CNN) architecture. Besides that, the framework operates by clustering reduced-dimensionality features which fails to explore information that non-linear relationships between the latent variables learned by the CNN model could provide. 


Following the same line of work employed by the authors of~\cite{feitoza2019hibrido, caron2019deeper}, we aim at combining these methods into a reformulated framework in the context of FGVC/intra-class variability mitigation. We reformulate the problem of self-supervised learning of visual features, as posed in the Deeper Cluster framework from~\cite{caron2019deeper}, to propose a new method for leveraging FGVC, based on the method of~\cite{feitoza2019hibrido}. Our method consists of a semi-supervised approach for learning fine-grained image features. We propose an end-to-end framework that integrates K-Means clustering, dimensionality reduction and hierarchical classification to leverage Fine-Grained Visual Categorization. Differently from~\cite{caron2019deeper}, we perform K-Means on every class, allowing the model to discover new class labels conditioned upon the supervised labels originally provided by a dataset. In turn, we then use these labels in a hierarchical classification process, in which the classification model is trained to predict both the parent (species) and subclass labels (K-means cluster assignments). 

The goal behind this work is to investigate whether the fine-grained features that we enforce a model to develop by training it to predict cluster assignments can be reused by the parent classifier to predict the target labels and improve the classification process as a whole. In order to do this, we conduct our experiments on the Pl@antNet300K dataset, which is a large-scale dataset for plant species identification, primarily used in fine-grained visual categorization (FGVC) tasks~\cite{plantnet300k}. The dataset includes images of plants taken from different viewpoints and parts (e.g., leaves, flowers, fruits), which increases the intra-class variability and complexity of the task -- specially considering the class imbalance --, making it an ideal candidate for testing the robustness of our approach. 

Our findings indicates that despite sharing the same backbone, features discovered in each classification level seem to be independent. Under this assumption, our findings suggests that because of this, parent classifiers can't benefit from information learned in lower levels in the hierarchy. Despite that, the method was still capable of matching the state-of-the-art performance on the dataset.

The remainder of this paper is organized as follows. Section 2 presents the related work and the core concepts that support this work. Section 3 provides a detailed description of the CNN architectures considered. Section 4 addresses the experimental evaluation process as well as the evaluation protocol and measures. In Sections 5 and 6 we present and discuss the results and conclusions, respectively


\section{Related Work}
\subsection{Intra-Class Variability and Deep Clustering}
Intra-class variability is a challenging adversity, that's mainly because when its images are way too dissimilar between themselves, a model can fail to uncover the feature patterns that represents a class~\cite{feitoza2019hibrido}. This can be especially harder on underrepresented classes in which there isn't enough data to allow the model to find generalizable patterns within it. In that sense, clustering algorithms could be a helpful tool that can leverage unsupervised learning to find groups of images with similar visual characteristics~\cite{caron2019deeper, caron2018deep}. Those algorithms could be applied to find sub-classes of images with greater cohesion in terms of visual similarity. In turn, these subgroups can be used as new classes in a supervised learning  (classification) process as those new classes may present more homogeneous characteristics, thereby simplifying the feature learning process.

The task of clustering involves grouping $N$ data samples into $K$ categories i.e., clusters. In the case of K-Means \cite{lloyd1982least}, this goal is attained by optimizing the cost:

\begin{equation}
    \min_{ M \in \R ^ {M \times K}, \{ s_i \in \R^K \} \hspace{0.15cm} s.t \hspace{0.15cm} s_i^{T}1 = 1} \sum_{i=1}^{N}  \| x_i - Ms_i  \|_{2}^{2} 
\end{equation}

Where $x_i \in \R \hspace{0.1cm} | \hspace{0.1cm} \{ x_i\}_{i=1,...,N}$, belongs to a set of data samples, $s_i$ corresponds to the assignment vector of data point $i$, such that $s_i \in \{0,1\} \hspace{0.1cm} \forall i$, that is, a binary vector with a single non-zero entry. Finally, $M$ corresponds to the centroid matrix, where each column corresponds to a centroid, and $K$ is the number of centroids. In the context of deep networks, the clustering process is resultant from the same cost optimization process, except that the set of data samples of which $x_i$ belongs, is generated by a nonlinear mapping i.e., $f_{\theta}(x_i)$ in Equation \ref{deep_clustering_2} -- where $\theta$ denotes the parameters of the feature extractor network backbone. 

\begin{equation}
    \min_{ M \in \R ^ {M \times K}, \{ s_i \in \R^K \} \hspace{0.15cm} s.t \hspace{0.15cm} s_i^{T}1 = 1} \sum_{i=1}^{N}  \| f_{\theta}(x_i) - Ms_i  \|_{2}^{2} 
    \label{deep_clustering_2}
\end{equation}

Besides that, clustering and DL can be jointly integrated by minimizing a cost function in the format:

\begin{equation}
    \min_{\theta, W} { \frac{1}{N} \sum_{i=1}^{N} l(z_i , f_{\theta, W}(x_i))} 
    \label{joint_deepcluster}
\end{equation}

Where $z_i$ corresponds to the target classification variable, $f_{\theta,W}(x_i)$ are the activations of the network backbone, followed by a linear classifier with parameters $W$, and $l$ is the corresponding loss function $l(\cdot;\cdot): \R^M \mapsto \R$ (e.g., crossentropy, a.k.a negative log-softmax). In the case of end-to-end Deep Clustering (DC), both approaches are integrated by obtaining $z_i=s_i$ (latent pseudo-label) via clustering, and optimizing by alternating between learning the parameters $\theta$ and $W$ and updating the pseudo-labels $z_i$~\cite{caron2019deeper}. On the other hand, in two-stage approaches features are extracted from a network backbone and to obtain pseudo-labels via clustering in a disjoint way. 

As we can see, the goal of assigning samples into distinct categories in a non-supervised fashion can by attained by this method. This way, K-Means can be employed for automatically discovering sub-classes, that is, image subsets of possible increased cohesion, in terms of visual similarity. These sub-classes in turn, can be employed as new classes in a distinct classification setting, in which the previous (e.g., species) target can be related to. As a result the model is now able to learn more fine-grained characteristics from the data and thereby attenuate the effects from intra-class variability, thus improving the target classification performance. 

In that sense, the authors from~\cite{feitoza2019hibrido} demonstrated that the training process involving sub-classes with greater cohesion enabled reducing the complexity of feature learning, in comparison to the classification process involving the original configuration of the classes. In their work, they investigated the potential of clustering methods (K-Means) to infer subgroups from species that presented images with significantly different visual characteristics in a plant species classification scenario. Despite of indubitable effectiveness improvement, when compared to end-to-end approaches, their work fails to leverage the advantages of DNNs in terms of learning capabilities, as in their framework feature learning and clustering are performed in distinct stages -- where on the other hand, more optimized features could be learned if the two approaches were combined. Besides that, feature extraction, dimensionality reduction and clustering were also employed separately. 

In contrast to~\cite{feitoza2019hibrido}, in the work of~\cite{caron2019deeper}, the authors proposed a framework that integrates deep clustering for pre-training CNN visual features without relying on annotations (self-supervised learning). Their method uses K-Means clustering to find pseudo-labels that are used to perform classification and feature learning without relying on data annotation. The author's framework integrates clustering, dimensionality reduction and feature learning in an end-to-end fashion. Their method operates by applying Principal Component Analysis (PCA) into the features generated by a CNN network and then clustering those features hierarchically with K-Means. After that, the K-Means cluster assignments are then employed in classification process, in which the CNN learns features by predicting both data augmentations and cluster assignments jointly. 

\subsection{Dimensionality Reduction}

Many learning frameworks treat dimensionality reduction (DR) and clustering separately, as opposed to optimizing the two tasks jointly, which can improve the performance of both~\cite{k_means_friendly}. Besides that, often times the models used to learn mappings from the higher-dimensional to low-dimensional latent spaces are linear, such as in the works of~\cite{feitoza2019hibrido, caron2019deeper, caron2018deep, dourado2021experimental}. The problem with this approach is that it fails to account for possible nonlinear relationships between the latent variables in a DNN model's output. Motivated by this, in~\cite{k_means_friendly}, the authors propose a method for learning "K-Means friendly" spaces. Their idea is that by employing DNNs to perform dimensionality reduction it can both account for nonlinearities and learn a more "cluster friendly" latent space, as clustering-promoting objectives could be explicitly incorporated in the learning process. 

This can be achieved by using an Autoencoder (AE) network \cite{regularized_auto_encoders} architecture, similarly trained to minimize a cost function as demonstrated below:

\begin{equation}
    \min_{\omega} { \frac{1}{N} \sum_{i=1}^{N} l( g_{\omega}(f_{\theta}(x_i))) , f_{\theta}(x_i))} 
    \label{autoencoder}
\end{equation}

In this case, the AE $g_{\omega}(x_i)$ is generally employed as a Multi-Layer Perceptron (MLP), a generative model composed by two stack of layer blocks: an encoder that sequentially reduces data dimensionality, and a decoder that projects the data back to the original input dimension -- where $\omega$ denotes the network parameters. The goal of this proposal is to reconstruct the input data (decoding) that was projected into a lower-dimensional latent space (encoding) through nonlinear transformations. Is expected that by training the model to reconstruct its own input it learns to uncover a lower dimensional manifold in which the data features lies in\cite{why_deep_learning_works}. In other words, it's expected that in order to to reconstruct its input from a lower dimension projection the model can learn a lower dimensional representation that can be more suitable to tasks that are more sensitive to high-dimensional data. The problem with this approach is that is prone to trivial solutions, i.e., identity mapping, outputting zeros~\cite{k_means_friendly}. Therefore, regularization mechanisms might have to be applied in order to prevent the model from collapsing.

This method allow to uncover possible nonlinear relationships between latent variables generated by DNNs while performing DR. Besides that, this framework can be integrated into a deep clustering pipeline to simultaneously cluster from reduced dimensions features, but with the additional possibility of integrating a clustering-promoting objective into the loss function $l$. In~\cite{k_means_friendly}, the authors formulate how this could be achieved by employing the AE network to perform the same cost optimization as in Equation~\ref{autoencoder}, but with a penalty term added to the loss function, which corresponds to the K-Means loss, sometimes known as \textit{intertia} and which is presented in Equaiton~\ref{k_means_friendly_loss}:

\begin{equation}
    \min_{\omega, M, \{s_i\} } { \frac{1}{N} \sum_{i=1}^{N} \biggl( l( g_{\omega}(f_{\theta}(x_i))) , f_{\theta}(x_i)) + \frac{\lambda}{2}{\| g_{\omega}^b(f_{\theta}(x_i)) - Ms_i  \|_{2}^{2}} \biggl)}
    \label{k_means_friendly_loss}
\end{equation}

Note that K-Means is performed over $g_{\omega}^b(f_{\theta}(x_i))$ rather than $g_{\omega}(f_{\theta}(x_i))$, that's because $g_{\omega}^b$ denotes the bottleneck layer output from the AE, which corresponds to the output of the encoder module. In this way, the authors from~\cite{k_means_friendly} incorporate the K-Means cost (L2 norm) into the reconstruction loss function depicted in Equation~\ref{k_means_friendly_loss}. This enables the model not only to generate reduced-dimensionality embeddings, but features that are more suitable to clustering processes. Although this process may not be straightforward, as $M, \theta$ and ${s_i}$ can not be optimized jointly via stochastic gradient descent (SGD) because ${s_i}$ is discrete~\cite{k_means_friendly}. Because of that, in~\cite{k_means_friendly}, the authors achieve this by computing the derivative of the K-Means cost with respect to the parameters of $f_{\theta}$: the DNN (e.g., vision transformer, CNN, etc) that generated features for dimensionality reduction.

\section{Proposed Method}\label{sec::pspotterk}


\begin{figure*}
    \centering
    \includegraphics[width=0.8\linewidth]{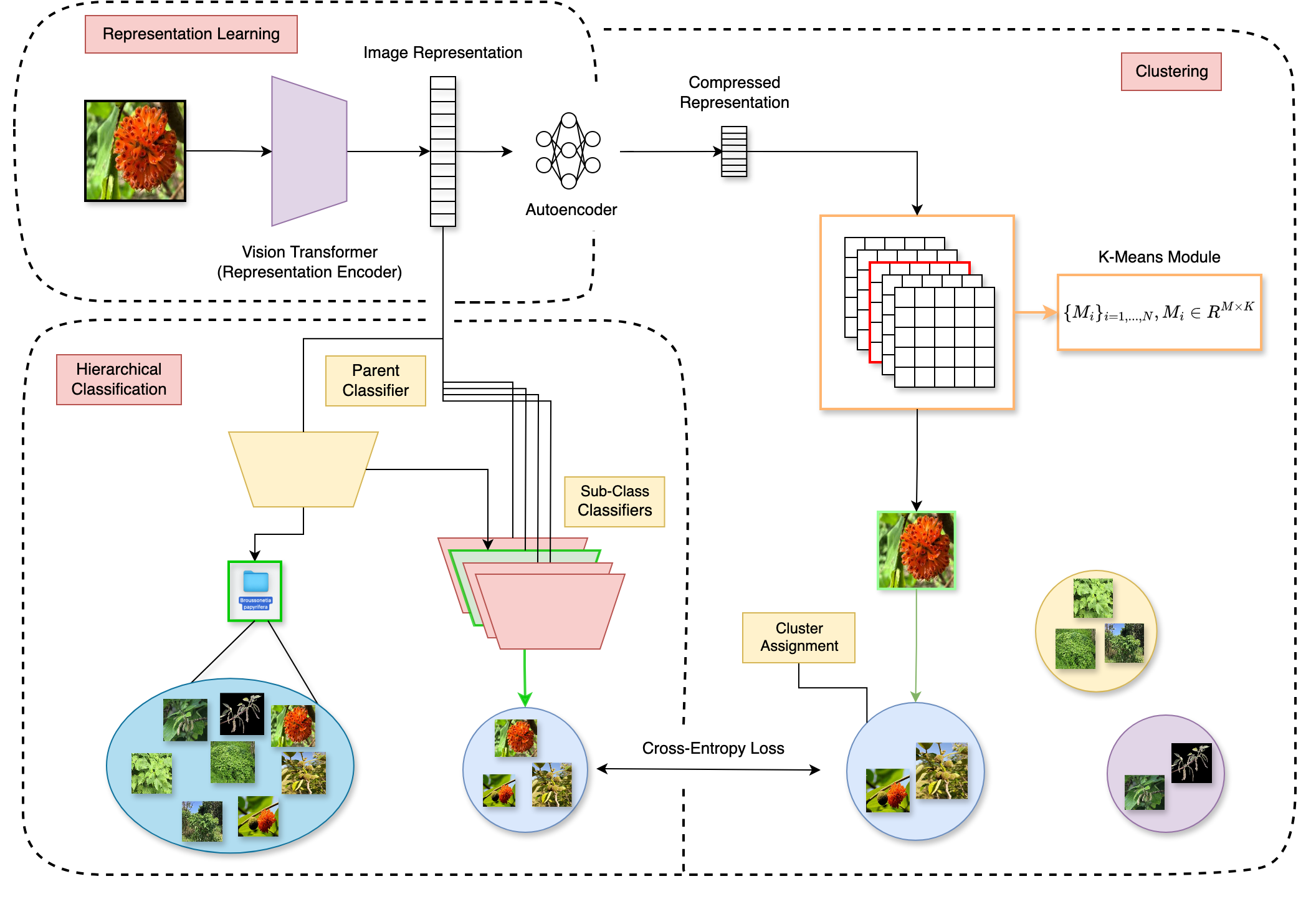}
    \caption{An overview of the Fine-Grained Deep Cluster Categorization Framework (FGDCC).}
    \label{simplified_view}
\end{figure*}

\begin{figure*}
  \includegraphics[width=\textwidth]{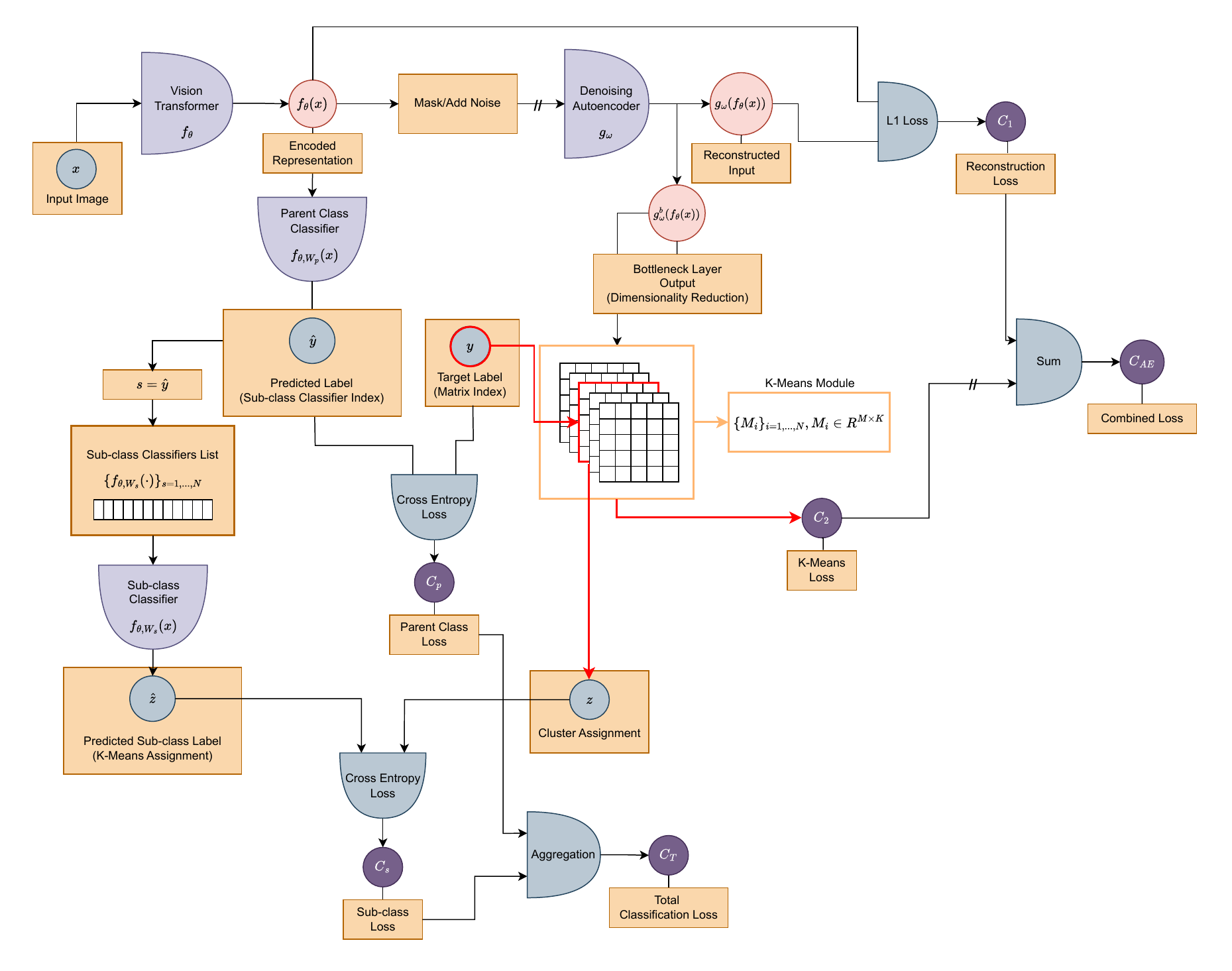}
  \caption{A detailed view on the FGDCC architecture.}
  \label{FGDCC}
\end{figure*}

In this work, we take as inspiration the ideas from~\cite{feitoza2019hibrido, caron2019deeper, k_means_friendly, caron2018deep} to propose a deep clustering framework focused on mitigating intra-class variability for FGVC tasks. Figure~\ref{simplified_view} presents a higher-level representation of our proposal. Similarly to~\cite{caron2019deeper}, we use K-Means to perform clustering over a  Vision Transformer~\cite{dosovitskiy2020image} (ViT) embeddings, and train a classifier on top of the ViT backbone to predict the cluster assignments. Differently from~\cite{caron2019deeper}, instead of clustering all the dataset into a fixed number ($K$) of sub-classes,
we perform class-wise clustering to discover sub-class labels that are employed in a hierarchical classification process -- species prediction followed by sub-class predictions. Besides that, instead of performing PCA, we train an autoencoder model to reduce feature dimensionality.  

A more detailed view on how our method is performed is presented in Figure~\ref{FGDCC}. Our method works as follows:  we assume that the target dataset is composed by a sequence of input images $\{ x_i\}_{i=1,...,N}$, each of which is associated to a label $y_n$ (target variable), of which we will interchangeably refer it as the parent class. In a similar way to~\cite{feitoza2019hibrido}, our method learns to discover sub-class labels $z_j$ by employing K-Means clustering over each subset of images from $\{ x_i\}$ from which a label $y_i$ is associated to. In other words, we employ K-Means over each parent class $y_n$ images, so as to find a set of labels $\{z_j\}_{j=1, ..., K}$ to which we construct an hierarchical relationship to $y_n$. As opposed to $y_n$ , $z_j$ is assumed to be composed by more visually similar images, because it represents the cluster assignment discovered via unsupervised aggregation of similar feature encodings. Therefore, $z_j$ encodes a latent degree of similarity between the feature vectors of the images that were assigned to it. 

We use a vision transformer~\cite{dosovitskiy2020image} as an encoding function $f_{\theta}(x_i)$, which allows a mapping between image space $\R^{D}$ into a feature space $\R^d$, where $d \ll D$, of which we reduce the dimensionality by training an autoencoder model $g_{\omega}(\cdot)$ to reconstruct the input $f_{\theta}(x_i) \in \R^d$, by employing the optimization demonstrated in Equation~\ref{autoencoder}. Then, we use the same autoencoder bottleneck layer $g_{\omega}^b(f_{\theta}(x_i)): \R^d \longmapsto \R^M$, $M < d$ to generate a more compatible representation to the K-Means model. On each dataset iteration, we encode a batch of images through the ViT encoder $f_{\theta}(x_i)$ which generates the latent representation used by the autoencoder to perform reconstruction (Equation~\ref{autoencoder}). The bottleneck layer output $g_{\omega}^b(f_{\theta}(x_i))$ goes into the K-Means module, that computes the cluster assignments for every image in the batch. In turn, the assignments are used as classification labels in a hierarchical classification process. In this step, we have a two-stage hierarchical classification module in which the parent classifier $f_{\theta, W_p}(x_i)$ takes as input the original feature encoding from the ViT and predicts the corresponding ground truth label $\hat{y}_n$, for each image in the batch. On the other hand, in order to predict the sub-class label (i.e., K-Means assignment), a sub-class classifier has to be selected among a list of sub-class classifiers $ C = \{f_{\theta, W_s}(\cdot)\}_{s=1, ..., N}$. This way, the parent classifier prediction $\hat{y}_n$ is used as index to select the sub-class classifier that also takes in as input the ViT backbone features to predict the corresponding cluster to which that image was assigned by the K-Means module. At the end of an epoch, the K-Means centroid matrices are updated as we have alternate between SGD and updating the cluster parameters because ${s_i}$ is constrained on a discrete set~\cite{k_means_friendly}. 

\subsection{Optimization and Clustering  Parameters Update}\label{joint_optimization}

During training, the ViT encoder backbone $f_{\theta}(\cdot)$ is hence jointly optimized for minimizing the following cost (hierarchical loss), where $l$ corresponds to the traditional cross-entropy loss function: 

\begin{equation}
    \min_{\theta, W_p, W_s} { \frac{1}{N} \sum_{i=1}^{N} \biggl( l(y_n , f_{\theta, W_p}(x_i)) + l(z_i , f_{\theta, W_s}(x_i))\biggl)} 
    \label{hierarchical_loss}
\end{equation}

Where $s = \hat{y}_i = f_{\theta, W_p}(x_i)$, and ${z_i}$ is obtained from the optimization process occurred at the end of each epoch, as  shown in Equation~\ref{k_means_with_autoencoder}:

\begin{equation}
        \min_{ M_i \in \R ^ {M \times K}, \{ z_i \in \R^K \} \hspace{0.15cm} s.t \hspace{0.15cm} z_i^{T}1 = 1} \sum_{i=1}^{N}  {\| g_{\omega}^b(f_{\theta}(x_i)) - M_iz_i  \|_{2}^{2}}
        \label{k_means_with_autoencoder}
\end{equation}

In turn, the autoencoder is optimized for reconstruction considering the cost depicted in Equation~\ref{k_means_friendly_loss}, where $l$ corresponds to a smoothed version of the L1 loss, as in~\cite{IJEPA}.

\subsection{Accounting for Model Selection}
For one global value of K i.e., in the advent of trying to discover a fixed number K of sub-classes for every parent class, the K-Means module corresponds to a set of K-Means matrices $\{M_i\}_{i=1, ..., N}, M_i \in \R^{M \times K}$, where $N$ corresponds to the number of dataset classes ($|\{y_n\}|$). Instead of assuming that all classes can be clustered in the same number of sub-classes, we perform a trick for handling model selection i.e., updating the classifier correspondent to the best K. First, we extend the number of sub-class classifiers, such that instead of having a one-to-one mapping between a parent and a sub-class classifier of output length K, now we will have a one-to-one mapping between a parent classifier and a list of sub-class classifiers. The optimization process now consists of assigning each image in the batch to each cluster within a set of clusters, ranging from $k=1, ..., K$. This way, for each image $i$ in the batch, instead of a single $z_i$, we now obtain a set of cluster assignments $\{z_k\}_{k=1, ..., K}^i$, from which we perform classification over the set of sub-class classifiers associated to the corresponding parent classifier $W_p$. In other words, as we associated a sub-class classifier for every $k \in K$, we choose to update only the parameters of the $k$ classifiers that achieved best classification performance across the batch -- that is, the ones that achieved the smallest cross-entropy error between the sub-class predictions ($K$ sub-class predictions) and the correspondent set of cluster assignments $\{z_k\}_{k=1, ..., K}^i$, for every image $i$ in the batch.



\subsection{Avoiding Trivial Solutions}
As posed in~\cite{caron2019deeper, k_means_friendly}, trivial solutions are expected for these models, specially when jointly optimized for a task. It is expected that when trained for reconstruction (Equation~\ref{autoencoder}), for example, autoencoders can suffer from several types of representation collapse. When combining together: classification over K-Means assignments to clusters generated by aggregation of features produced by an autoencoder, the system becomes prone to even more trivial solutions. In the first case, in order to circumvent possible trivial solutions, we train a denoising autoencoder with additional regularization strategies as suggested in~\cite{k_means_friendly}. In the latter, trivial solutions can happen in the sense that the K-Means models could learn to assign all elements to a single cluster, which thereby facilitates the learning process for the classifiers, preventing it from learning meaningful representations. Therefore we employ the solution proposed by~\cite{caron2018deep}, which consists of replacing empty clusters by non-empty ones with small perturbations.

\subsection{Object of Study}
In order to assess the impact of the proposed method in the context of FGVC, we evaluated the species classification performance in the Pl@ntNet300K dataset~\cite{plantnet300k}. The Pl@ntNet300K dataset is a large-scale fine-grained visual categorization dataset for plant species classification. It consists of 306,146 images taken from various viewpoints and plant organs e.g., leaves, flowers, fruits belonging to 1081 classes (plant species). Despite the long-tail distribution/class-imbalance (about 80\% of the classes composes 11\% of the data), label ambiguity, intra-class variability and other issues~\cite{plantnet300k}, the Pl@antNet300K is a widely used dataset in the research community. The train dataset is consisted of 243,916 images, whereas validation and test datasets has 31,118 and 31,112 images respectively. As we identified, there were cases of species with less than 5 samples, which would therefore make impossible to perform clustering (e.g., with say, $K=5$). Because of that, we performed upsampling in the train dataset to 10 samples all of such classes. To do this we applied random crop augmentations, this increased the train partition by 1983 images. 

\subsection{Experimental Settings}
This section presents the experimental evaluation settings, such as neural network architectures, hyperparameters, experiment configurations, etc. For the Vision Transformer~\cite{dosovitskiy2020image}, we use the ViT-Huge implementation and pre-trained  ImageNet-22k weights from the I-JEPA~\cite{IJEPA} authors. For the autoencoder architecture, we evaluated a symmetric encoder-decoder architecture consisting of linear layers followed by Gaussian Error Linear Units activations (GELU)~\cite{gelu}. Two bottleneck outputs length were evaluated: 256 and 384 dimensions. In the first case the encoder block consists of 4 linear layers that gradually reduces the original input dimensionality from 1280 into 256 dimensions in steps of 256. In the second case, it consists of 3 linear layers where the reduction occurs in steps of 256, whereas in the bottleneck layer it decreases the dimension by a half, to output a 384 dimensions representation. 

To account for clustering, we used the Faiss library~\cite{faisscpu, faissgpu} to auxiliate our implementations. In this case, the clustering module consisted of a set of Faiss K-Means objects for each class. More precisely, we evaluated K-Means with $k=\{2,3,4,5\}$, which, in this case corresponded to associating a list of Faiss K-Means objects for each value of $K \in k$ to each class. Although in~\cite{caron2019deeper} and~\cite{k_means_friendly} cluster parameters update occurs on the basis of $T > 1$ epochs, we evaluate updating it at the end of every epoch.      

The whole system was trained for 50 epochs over a single Nvidia 80G A-100 GPU with a batch size of 96 images and two independent AdamW~\cite{zhou2024towards} optimizers for the ViT and AE models. For the ViT-H fine-tuning, we extract the embeddings after averaging over the patch-level representation of the model's last block and adopt the fine-tuning recipe from~\cite{he2022masked}, as recommended by the authors~\cite{IJEPA}. We don't apply Mixup~\cite{zhang2017mixup} or CutMix~\cite{yun2019cutmix}. For the ViT-H we used a Cosine Learning Rate Decay Scheduler~\cite{IJEPA}, with start\_lr=$7.5e^{-5}$, lr= $2.5e^{-4}$ and final\_lr=$1e^{-6}$. The AE was trained with a default learning rate of $1e^{-3}$, standard weight decay and gradient clipping with gradient norm=1.0. For performance evaluation and comparison we report the Top-k (Top-1 and Top-5) test set accuracy results as in~\cite{plantnet300k}.

\section{Results and Discussion}
\begin{figure}[t]
    \includegraphics[width=\columnwidth]{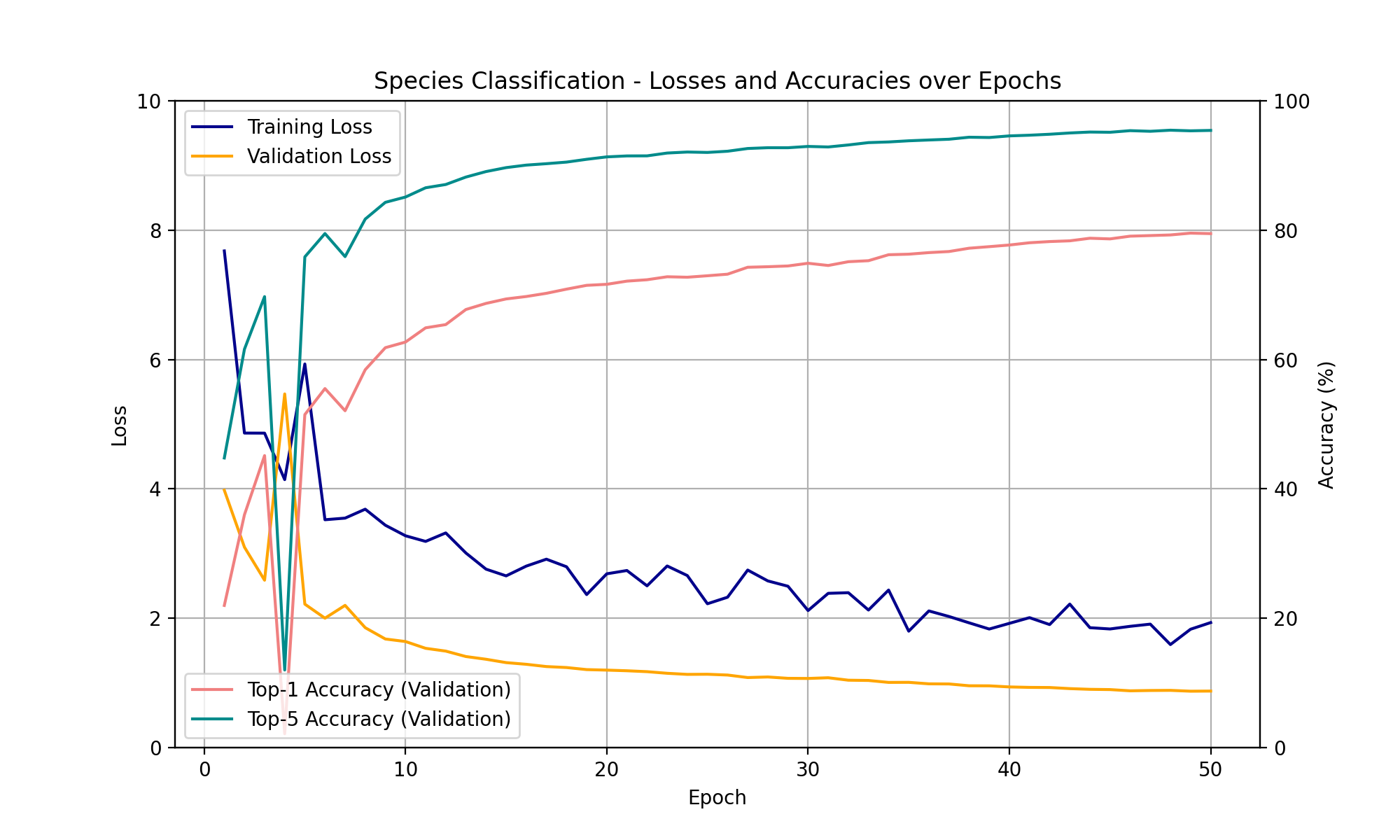}
    \caption{Species classification training and validation statistics over the epochs.}
    \label{species_stats}
\end{figure}

In this section, we present the results from experiments conducted in the Pl@antNet300K dataset for the FGDCC architecture. In Figure~\ref{species_stats}, we observe the results obtained for Species classification. Despite presenting a tendency towards convergence, it's not possible to claim that the model was inevitably converging. That's because we used a cosine scheduler~\cite{IJEPA} that performs learning rate decay over the epochs and could possibly have induced a convergence behavior. Despite that, the model matches the state-of-the-art test-set performance posed in~\cite{plantnet300k}, with Top-1 and Top-5 accuracies of respectively 79.756\% and 96.324\%.

\begin{figure}[t]
    \centering
    \includegraphics[width=0.5\textwidth]{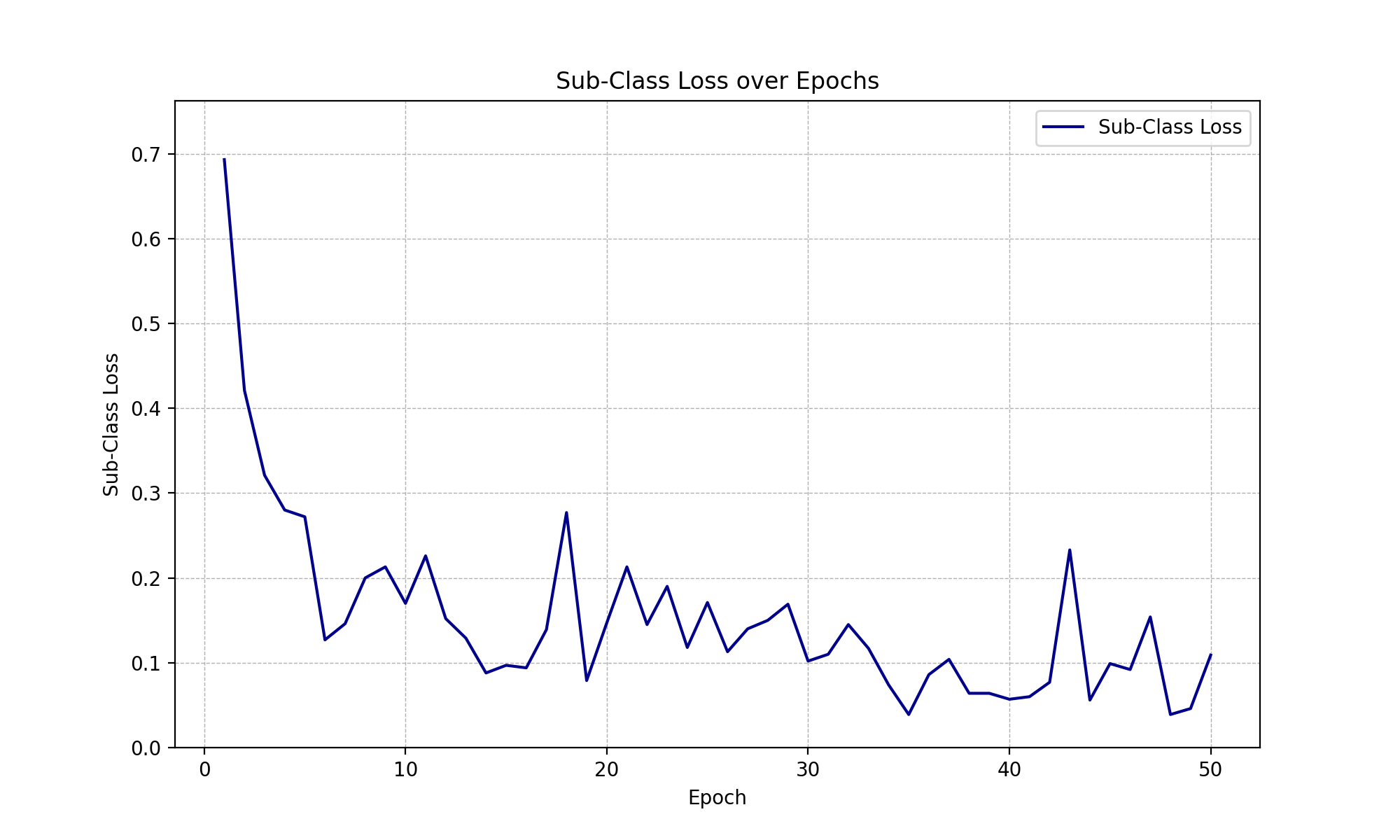}
    \caption{System sub-class classification behavior per epoch - average across the best K values per species. }
    \label{sub_class_loss}
\end{figure}

Figure~\ref{sub_class_loss} demonstrates the average sub-class loss across the epochs i.e., cross-entropy between sub-class classifiers (best classifiers) predictions and K-Means cluster assignments. As we can see, it indicates that there were subparts of the system that could be optimized even further. Despite that, its hard to say to which extent this loss can converge in a smooth way, as it also depends on another components that are optimized independently. In other words, some variance in this loss function will be always expected due to (1) - Mistakes over the parent classifier i.e., leading to the wrong selection of the sub-class classifier; (2) - Sensitivity of the loss function to the changes in the sub-class labels that are modified upon the optimization of the K-Means centroids, which in turn also depend on the optimization of the autoencoder and thereby the features that it provides; (3) - Model selection: the system is trained to optimize the best performing K-classifier for each class, therefore variance in the loss function can occur in this sense as well, as stochastic optimization plays a significant role. Despite that, in an overall way, the sub-class loss demonstrates a descending behavior which indicates that optimization is tending towards a point of minima for those experimental settings.    

\begin{figure}[t]
    \centering
    \includegraphics[width=0.5\textwidth]{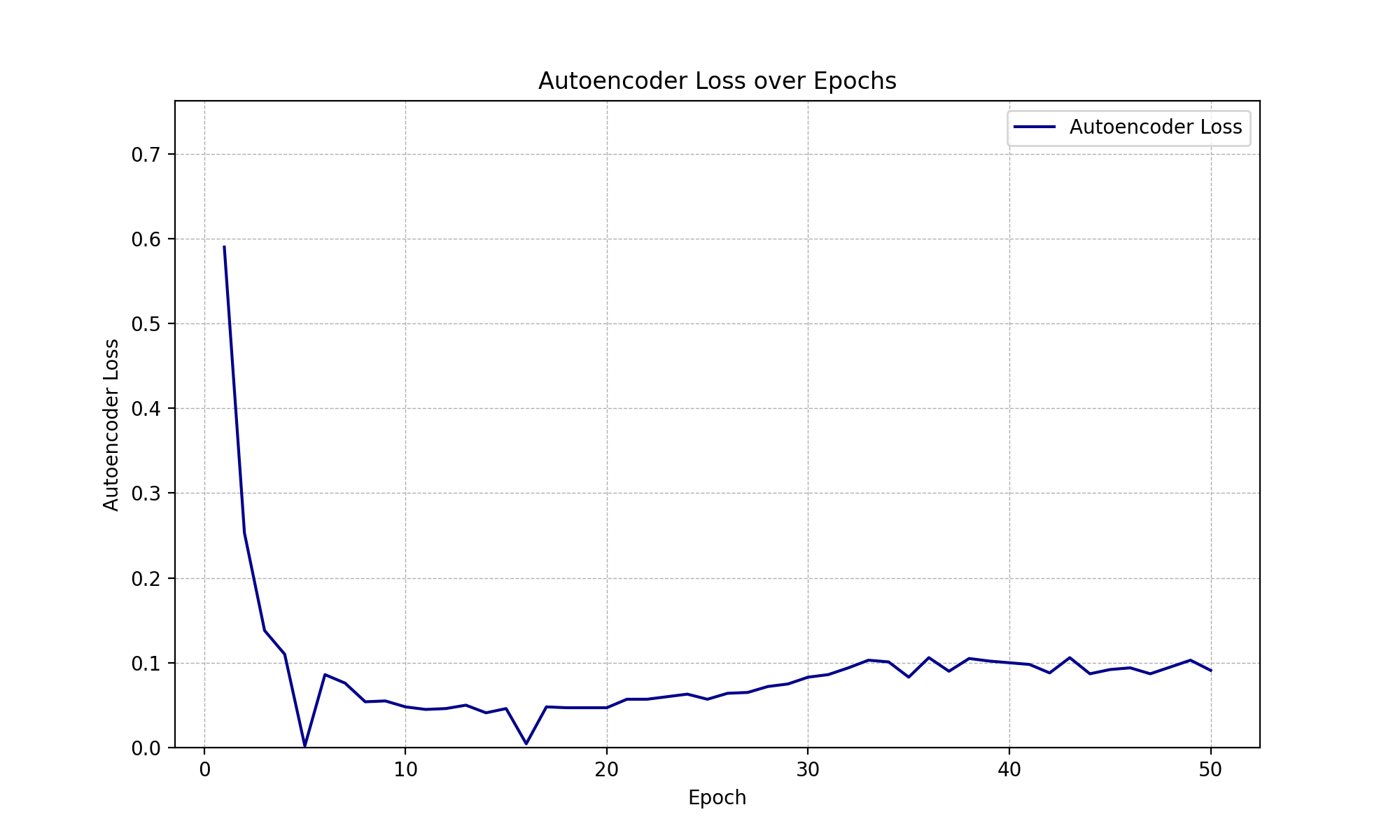}
    \caption{Autoencoder reconstruction loss across the epochs.}
    \label{autoencoder_loss}
\end{figure}

Figure~\ref{autoencoder_loss}, in turn demonstrates the autoencoder's L1 loss graph. We can see that the loss decreases steadily, reaching its lower points around epochs 5 and 17 then increasing to a point of convergence until epoch 50. This slightly ascending behavior is also expected since the ViT representation is changing through the course of the epochs which is also making the reconstruction harder as the ViT features becomes more semantically aligned with the classification task. 

\begin{figure}[t]
    \centering
    \includegraphics[width=0.5\textwidth]{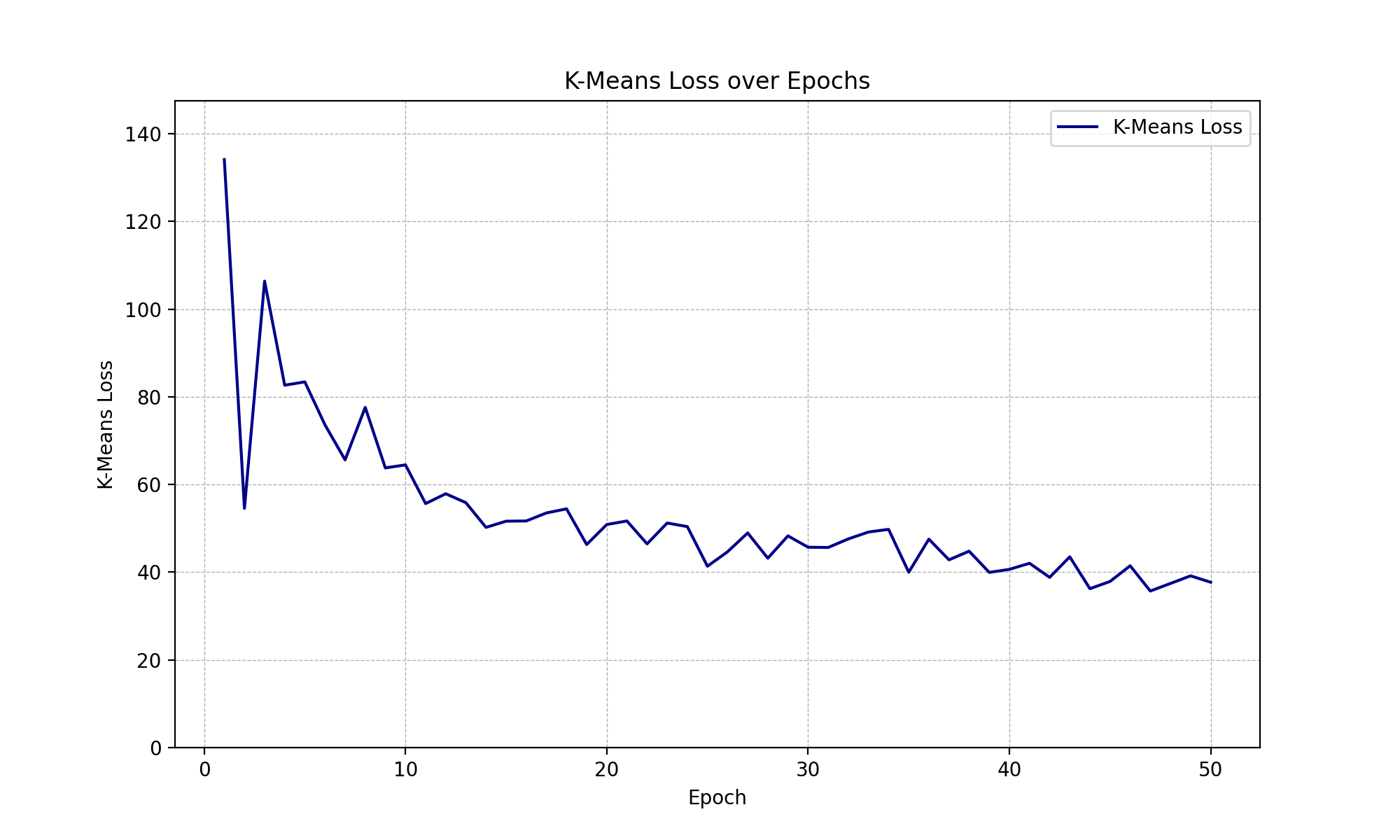}
    \caption{K-Means loss/Inertia across the epochs.}
    \label{k_means_loss_figure}
\end{figure}

Finally, in Figure~\ref{k_means_loss_figure}, a similar behavior to the sub-class loss occurs, in the sense that training for a few more epochs could enable to find more insights about the behavior of the model. Besides that, it seems to decrease more steadily in comparison to the sub-class loss, yet, it still demonstrates some degree of variance, which is also somewhat expected, as its optimization depends on the quality of the autoencoder features that in turn suffers the effect of the modification of the ViT features across the training epochs. 

In general, we observe that despite matching the state-of-the-art in terms of validation and test accuracies for species classification, there were some parts of the system that could be optimized even further, perhaps leading to more promising results. This is somewhat expected, as the system has components that are independently optimized. The architecture also has several keypoints that makes it sensible to cold-start, which in turn makes sense that it possibly needs more training epochs in relation to training a Vision Transformer in a standard way. From our perspective, the evidence seen so far contributes to the hypothesis that the system's ViT backbone is being trained to generate features that are invariant to classification level in the classification hierarchy. Because of that, the features learned by the sub-class classifiers doesn't seem impact directly over the features used by the parent classifier i.e., information is currently not being shared. We can observe this from looking into the parent classifier optimization behavior in Figure~\ref{species_stats}. We observed that the training behavior was very likely to the training behavior of a standard vision transformer model, as demonstrated by~\cite{plantnet300k}. Besides that, training behavior is the same regardless of clustering regularization. In other words, training with the regularization strategy proposed by~\cite{caron2018deep} generated the same behavior as training without it. This reinforces the claim that information doesn't seem to be shared across the classification hierarchy (at least not from lower levels into higher ones), which, if the opposite was true, it would be expected to see a different behavior for better or worse -- which doesn't seem to be the case here. 

Besides that, we believe that there are at least four main points of investigation to address in future experiments, each one regarding one subpart of the system. The first aspect regards empty clusters regularization: we believe that monitoring the number of empty clusters per epoch and the average number of empty clusters per $K$ value perhaps can give us more information regarding how effectively the solution proposed by~\cite{caron2018deep} works. The second aspect concerns information sharing between classifiers, i.e., it may involve testing classification strategies other than hierarchically. The third aspect regards model selection criteria: we believe that testing other K selection criteria such as~\cite{cosine_clustering_index} may be necessary for ablation purposes. Finally, regarding the autoencoder model, one perhaps critical aspect was not taken into consideration: we trained the autoencoder to minimize the mean absolute difference (L1 norm) between reconstructed features and the original (ViT backbone) features. As a consequence, this criterion makes representations more robust to outliers -- which perhaps is better suited for long-tailed distribution FGVC tasks --, as the penalty for large errors is linear and it encourages sparsity in learned representations. However, this sparsity constraint might not align well with the K-Means assumptions that clusters are spherical and equally sized in Euclidean space (L2 norm smoothness)~\cite{lloyd1982least, macqueen1967some}. This representation misalignment may be causing issues (e.g., suboptimal clustering performance), as the assumption of spherical clusters is violated when many features are zero and the actual distribution of points may become non-spherical. Besides that, sparse features can lead to clusters of varying densities and sizes, which K-Means is not well-equipped to handle due to its assumption of equally sized clusters. Therefore, testing other reconstructions such as L2 norm is mandatory in order to obtain more conclusions.




\section{Conclusion}
In this work, we presented FGDCC: Fine-Grained Deep Cluster Categorization, an architecture for intra-class variability problems in FGVC tasks. We evaluate the proposed method in a consolidated dataset, widely known in the research community for its FGVC context. The experiments conducted so far enabled an analysis in terms of convergence properties, overall (species-level) classification performance and optimization interoperability behavior between the system components. In general terms, the model demonstrated to match the state-of-the-art performance. Despite that, not enough conclusive evidence demonstrated that the system can improve parent class classification -- an thereby mitigate the effects of intra-class variability --  by performing class-wise cluster assignment categorization. Nevertheless, the experiments demonstrated that there are still several points of improvement and experiments to be conducted in order to obtain more conclusive evidence. Because of that we firmly believe that future work will enable to demonstrate that the architecture proposed is in fact promising. In that sense, it was possible to verify that the model could be trained for more epochs in order to see its behavior in terms of longer training routines. Besides that, several modifications to the autoencoder training routines, learning rate scheduling, regularization strategies and architectural constraints could be evaluated to verify the consequences over the overall system behavior. Finally, if verified that hierarchical classification doesn't enable to effectively learn features that can be somehow reused by the parent classifier, future work can be performed in the sense of testing another classification approaches, such as making predictions over the space of the cartesian-product between the K-Means assignments and the dataset labels~\cite{caron2019deeper}. Besides that, different performance metrics will also have to be accounted in order to make proper assessments, especially considering the long-tailed distribution and therefore class imbalance context.     

\ifCLASSOPTIONcaptionsoff
  \newpage
\fi

\bibliographystyle{IEEEtran}
\bibliography{IEEEabrv.bib,contents/referencias.bib}

\end{document}